\documentstyle[colacl,11pt]{article}

\title{{\bf Applying System Combination to Base Noun Phrase Identification}}

\author{
Erik F. Tjong Kim Sang$^\alpha$,
Walter Daelemans$^\alpha$,
Herv\'e D\'ejean$^\tau$,\\
{\bf Rob Koeling$^\gamma$,
Yuval Krymolowski$^\beta$,
Vasin Punyakanok$^\iota$,
Dan Roth$^\iota$}\\\\
\begin{tabular}{ccc}
$^\alpha$University of Antwerp & 
$^\tau$Universit\"{a}t T\"{u}bingen & 
$^\gamma$SRI Cambridge\\
Universiteitsplein 1 & Kleine Wilhelmstra{\ss}e 113 & 
                                                   23 Millers Yard,Mill Lane\\
B-2610 Wilrijk, Belgium & D-72074 T\"ubingen, Germany & Cambridge, CB2 1RQ, UK\\
\{erikt,daelem\}@uia.ua.ac.be & dejean@sfs.nphil.uni-tuebingen.de &
koeling@cam.sri.com
\end{tabular}\\\\
\begin{tabular}{cc}
$^\beta$Bar-Ilan University & $^\iota$University of Illinois\\
Ramat Gan, 52900, Israel & 1304 W. Springfield Ave.\\
yuvalk@macs.biu.ac.il & Urbana, IL 61801, USA\\
                      & \{punyakan,danr\}@cs.uiuc.edu
\end{tabular}
}

\date{\today}

\begin{document}

\maketitle

{
\begin{picture}(0,0)
\put(0,220){
\makebox(440,0){
In:
{\em Proceedings of COLING 2000},
pages 857--863,
Saarbr\"ucken, Germany, 2000.
}}
\end{picture}
}

\vspace*{-0.9cm}
\begin{abstract}
\noindent
We use seven machine learning algorithms for one task:
identifying base noun phrases.
The results have been processed by different system combination 
methods and all of these outperformed the best individual result.
We have applied the seven learners with the best combinator, a
majority vote of the top five systems, to a standard data set and managed 
to improve the best published result for this data set.
\end{abstract}

\section{Introduction}

Van Halteren et al. \shortcite{vanhalteren98} and 
Brill and Wu \shortcite{brill98} 
show that part-of-speech
tagger performance can be improved by combining different taggers.
By using techniques such as majority voting, errors made by the
minority of the taggers can be removed.
Van Halteren et al. \shortcite{vanhalteren98} 
report that the results of such a combined
approach can improve upon the accuracy error of the best individual
system with as much as 19\%.
The positive effect of system combination for non-language processing
tasks has been shown in a large body of machine learning work.

In this paper we will use system combination for identifying
base noun phrases (baseNPs).
We will apply seven machine learning algorithms to the 
same baseNP task.
At two points we will apply combination methods.
We will start with making the systems process five output
representations and combine the results by choosing the majority 
of the output features.
Three of the seven systems use this approach.
After this we will make an overall combination of the results of 
the seven systems.
There we will evaluate several system combination methods.
The best performing method will be applied to a standard data set for
baseNP identification.

\section{Methods and experiments}

In this section we will describe our learning task: recognizing
base noun phrases.
After this we will describe the data representations we used and
the machine learning algorithms that we will apply to the task.
We will conclude with an overview of the combination methods that
we will test.

\subsection{Task description}
\label{sec-task}

Base noun phrases (baseNPs) are noun phrases which do not contain
another noun phrase.
For example, the sentence 

\begin{quote}
In $[$ early trading $]$ in $[$ Hong Kong $]$\\
$[$ Monday $]$ , $[$ gold $]$ was quoted at\\
$[$ \$ 366.50 $]$ $[$ an ounce $]$ .
\end{quote}

\noindent
contains six baseNPs (marked as phrases between square brackets).
The phrase {\it \$ 366.50 an ounce} is a noun phrase as well.
However, it is not a baseNP since it contains two other noun phrases. 
Two baseNP data sets have been put forward by Ramshaw and Marcus 
\shortcite{ramshaw95}.
The main data set consist of four sections of the Wall Street Journal
(WSJ) part of the Penn Treebank \cite{marcus93} as training material 
(sections 15-18, 211727 tokens) and one
section as test material (section 20, 47377 tokens)\footnote{This 
Ramshaw and Marcus \shortcite{ramshaw95} baseNP data
set is available via ftp://ftp.cis.upenn.edu/pub/chunker/}. 
The data contains words, their part-of-speech (POS) tags as computed
by the Brill tagger and their baseNP segmentation as derived from the
Treebank (with some modifications).

In the baseNP identification task, performance is measured with three 
rates.
First, with the percentage of detected noun phrases that are correct
(precision). 
Second, with the percentage of noun phrases in the data that were
found by the classifier (recall).
And third, with the F$_{\beta=1}$ rate which is equal to
(2*precision*recall)/(precision+recall).
The latter rate has been used as the target for optimization.

\subsection{Data representation}
\label{sec-data-repr}

In our example sentence in section \ref{sec-task}, noun phrases are
represented by bracket structures.
It has been shown by Mu\~noz et al. \shortcite{munoz99} that for 
baseNP recognition, 
the representation with brackets outperforms other data
representations. 
One classifier can be trained to recognize open brackets (O) and 
another can handle close brackets (C).
Their results can be combined by making pairs of open and
close brackets with large probability scores.
We have used this bracket representation (O+C) as well.
However, we have not used the combination strategy from 
Mu\~noz et al. \shortcite{munoz99}
but instead used the strategy outlined in Tjong Kim Sang 
\shortcite{tks2000a}:
regard only the shortest possible phrases between candidate open 
and close brackets as base noun phrases.

An alternative representation for baseNPs has been put forward by
Ramshaw and Marcus \shortcite{ramshaw95}.
They have defined baseNP recognition as a tagging task:
words can be inside a baseNP (I) or outside a baseNP (O).
In the case that one baseNP immediately follows another baseNP,
the first word in the second baseNP receives tag B.
Example:

\begin{quote}
In$_O$
early$_I$
trading$_I$
in$_O$
Hong$_I$
Kong$_I$
Monday$_B$
,$_O$
gold$_I$
was$_O$
quoted$_O$
at$_O$
\$$_I$
366.50$_I$
an$_B$
ounce$_I$
.$_O$
\end{quote}

\noindent
This set of three tags is sufficient for encoding baseNP structures
since these structures are nonrecursive and nonoverlapping.

Tjong Kim Sang \shortcite{tks2000a} 
outlines alternative versions of this tagging representation.
First, the B tag can be used for the first word of every baseNP
(IOB2 representation).
Second, instead of the B tag an E tag can be used to mark the last
word of a baseNP immediately before another baseNP (IOE1).
And third, the E tag can be used for every noun phrase final word
(IOE2).
He used the Ramshaw and Marcus \shortcite{ramshaw95} representation 
as well (IOB1).
We will use these four tagging representations and the O+C
representation for the system-internal combination experiments.

\subsection{Machine learning algorithms}

This section contains a brief description of the seven machine learning
algorithms that we will apply to the baseNP identification task:
ALLiS, {\sc c5.0}, IGTree, MaxEnt, MBL, MBSL and SNoW.

ALLiS\footnote{A demo of the NP and VP chunker is available at
http://www.sfb441.uni\-tuebingen.de/\~{ }dejean/chunker.h tml } 
(Architecture for Learning Linguistic Structures) is a learning system
which uses theory refinement in order to learn non-recursive NP and VP
structures
\cite{dejean2000a}. 
ALLiS generates a regular expression grammar which describes the phrase
structure (NP or VP).
This grammar is then used by the CASS parser \cite{AbneyCass}.
Following the principle of theory refinement, the learning task is composed of
two steps.
The first step is the generation of an \textit{initial grammar}.
The generation of this grammar uses the notion of default values and some
background knowledge which provides general expectations concerning the inner
structure of NPs and VPs.
This initial grammar provides an incomplete and/or incorrect analysis of the
data. 
The second step is the refinement of this grammar.
During this step, the validity of the rules of the initial grammar is checked
and the rules are improved (refined) if necessary.
This refinement relies on the use of two operations: the
contextualization (in which contexts such a tag always belongs to the
phrase) and lexicalization (use of information about the words and not
only about POS).

{\sc c5.0}\footnote{Available from http://www.rulequest.com}, a
commercial version of {\sc c4.5} \cite{Quinlan93}, performs {\em
top-down induction of decision trees}\/ ({\sc tdidt}). On the basis of
an instance base of examples, {\sc c5.0} constructs a decision tree
which compresses the classification information in the instance base
by exploiting differences in relative importance of different
features. Instances are stored in the tree as paths of connected nodes
ending in leaves which contain classification information. Nodes are
connected via arcs denoting feature values.  Feature information gain
(mutual information between features and class) is used to determine
the order in which features are employed as tests at all levels of the
tree \cite{Quinlan93}.
With the full input representation (words and POS tags), we were not
able to run complete experiments. We therefore experimented only with
the POS tags (with a context of two left and right). 
We have used the default parameter setting with decision trees
combined with value grouping.

We have used a nearest neighbor algorithm ({\sc ib1-ig}, here listed
as MBL) and a decision tree algorithm (IGTree) from the TiMBL learning
package \cite{timbl99}. 
Both algorithms store the training data and classify new items
by choosing the most frequent classification among training items 
which are closest to this new item. 
Data items are represented as sets of feature-value pairs.
Each feature receives a weight which is based on the
amount of information which it provides for computing the
classification of the items in the training data.
{\sc ib1-ig} uses these weights for computing the distance between a
pair of data items and IGTree uses them for deciding which 
feature-value decisions should be made in the top nodes of 
the decision tree \cite{timbl99}.
We will use their default parameters except for the {\sc ib1-ig}
parameter for the number of examined nearest neighbors (k) which we
have set to 3
\cite{daelemans99}.  
The classifiers use a left and right context of four words and
part-of-speech tags.
For the four IO representations we have used a second processing stage
which used a smaller context but which included information about
the IO tags predicted by the first processing phase
\cite{tks2000a}.

When building a classifier, one must gather evidence for predicting
the correct class of an item from its context. The Maximum Entropy
(MaxEnt) framework is especially suited for integrating evidence from
various information sources. Frequencies of evidence/class combinations
(called features) are extracted from a sample corpus and considered to
be properties of the classification process. Attention is constrained
to models with these properties. The MaxEnt principle now demands that
among all the probability distributions that obey these constraints,
the most uniform is chosen. During training, features are assigned
weights in such a way that, given the MaxEnt principle, the training
data is matched as well as possible. During evaluation it is tested
which features are {\em active} (i.e. a feature is active when the
context meets the requirements given by the feature). For every class
the weights of the active features are combined and the best scoring
class is chosen \cite{bdp96}. For the classifier built here the
surrounding words, their POS tags and baseNP tags predicted for the
previous words are used as evidence. A mixture of simple features
(consisting of one of the mentioned information sources) and complex
features (combinations thereof) were used. The left context never
exceeded 3 words, the right context was maximally 2 words. The model
was calculated using existing software \cite{Dehaspe97}.

\begin{table*}[t]
\begin{center}
\begin{tabular}{|l|c|c|c|c|c|c|c|c|c|}\cline{2-10}
\multicolumn{1}{l|}{} & 
\multicolumn{3}{c|}{MBL} &
\multicolumn{3}{c|}{MaxEnt} &
\multicolumn{3}{c|}{IGTree} \\\cline{2-10}
\multicolumn{1}{l|}{section 21}&O&C&F$_{\beta=1}$&O&C&F$_{\beta=1}$&O&C&F$_{\beta=1}$\\\hline
IOB1    &97.81\%&97.97\%&91.68&97.90\%&98.11\%&92.43&96.62\%&96.89\%&87.88\\
IOB2    &97.63\%&97.96\%&91.79&97.81\%&98.14\%&92.14&97.27\%&97.30\%&90.03\\
IOE1    &97.80\%&97.92\%&91.54&97.88\%&98.12\%&92.37&95.88\%&96.01\%&82.80\\
IOE2    &97.72\%&97.94\%&92.06&97.84\%&98.12\%&92.13&97.19\%&97.62\%&89.98\\
O+C     &97.72\%&98.04\%&92.03&97.82\%&98.15\%&92.26&96.89\%&97.49\%&89.37\\\hline
Majority&98.04\%&98.20\%&92.82&97.94\%&98.24\%&92.60&97.70\%&97.99\%&91.92\\\hline
\end{tabular}
\end{center}
\caption{The effects of system-internal combination by using different
output representations.
A straight-forward majority vote of the output yields better bracket
accuracies and F$_{\beta=1}$ rates than any included individual classifier.
The bracket accuracies in the columns O and C show what percentage of
words was correctly classified as baseNP start, baseNP end or neither.
}
\label{tab-res-internal}
\end{table*}

MBSL \cite{argamon99} uses POS data in order to identify baseNPs.
Inference relies on a memory which contains all the occurrences
of POS sequences which appear in the beginning, or the end, of a
baseNP (including complete phrases). These sequences may include a few
context tags, up to a pre-specified {\em max\_context}.
During inference, MBSL tries to 'tile' each POS string with parts
of noun-phrases from the memory. If the string could be fully covered
by the tiles, it becomes part of a candidate list, ambiguities
between candidates are resolved by a constraint propagation algorithm.
Adding a context extends the possibilities for tiling, thereby
giving more opportunities to better candidates.
The approach of MBSL to the problem of identifying baseNPs is
sequence-based rather than word-based, that is,
decisions are taken per POS sequence, or per candidate, but not for
a single word. In addition, the tiling process gives no preference to
any direction in the sentence. The tiles may be of any length, up to
the maximal length of a phrase in the training data, which gives MBSL
a generalization power that compensates for the setup of using only
POS tags.
The results presented here were obtained by optimizing MBSL parameters
based on 5-fold CV on the training data.

SNoW uses the Open/Close model, described in Mu\~noz et al.
\shortcite{munoz99}.
As is shown there, this model produced better results than the
other paradigm evaluated there, the Inside/Outside paradigm.
The Open/Close model consists of two SNoW predictors, one of which
predicts the beginning of baseNPs (Open predictor), and the other
predicts the end of the phrase (Close predictor).  The Open
predictor is learned using SNoW \cite{CCRR99,Roth98} as a function
of features that utilize words and POS tags in the sentence and,
given a new sentence, will predict for each word whether it is the
first word in the phrase or not. For each Open, the Close
predictor is learned using SNoW as a function of features that
utilize the words in the sentence, the POS tags and the open
prediction. It will predict, for each word, whether it can be the
end of the phrase, given the previously predicted Open. Each pair
of predicted Open and Close forms a candidate of a baseNP. These
candidates may conflict due to overlapping; at this stage, a
graph-based constraint satisfaction algorithm that uses the
confidence values SNoW associates with its predictions is
employed. This algorithm ("the combinator") produces the list of
the final baseNPs for each sentence.  Details of SNoW, its
application in shallow parsing and the combinator's algorithm are
in Mu\~noz et al. \shortcite{munoz99}.

\subsection{Combination techniques}
\label{sec-combi}

At two points in our noun phrase recognition process we will use
system combination.
We will start with system-internal combination: apply the same
learning algorithm to variants of the task and combine the results.
The approach we have chosen here is the same as in
Tjong Kim Sang \shortcite{tks2000a}:
generate different variants of the task by using different
representations of the output (IOB1, IOB2, IOE1, IOE2 and O+C).
The five outputs will converted to the open bracket representation (O) 
and the close bracket representation (C) and after this, the most 
frequent of the five analyses of each word will chosen
(majority voting, see below).
We expect the systems which use this combination phase to perform
better than their individual members \cite{tks2000a}.

Our seven learners will generate different classifications of the
training data and we need to find out which combination techniques 
are most appropriate.
For the system-external combination experiment, we have evaluated
different voting mechanisms, effectively the voting methods as
described in Van Halteren et al. \shortcite{vanhalteren98}.
In the first method each classification receives the same weight and 
the most frequent classification is chosen (Majority).
The second method regards as the weight of each individual
classification algorithm its accuracy on some part of the
data, the tuning data (TotPrecision). 
The third voting method computes the precision of each assigned tag
per classifier and uses this value as a weight for the classifier in
those cases that it chooses the tag (TagPrecision).
The fourth method uses both the precision of each assigned tag and
the recall of the competing tags (Precision-Recall).
Finally, the fifth method uses not only a weight for the current
classification but it also computes weights for other possible 
classifications.
The other classifications are determined by examining the tuning data
and registering the correct values for every pair of classifier results 
(pair-wise voting, see Van Halteren et al. \shortcite{vanhalteren98}
for an elaborate explanation).

Apart from these five voting methods we have also processed the
output streams with two classifiers: MBL and
IGTree.
This approach is called classifier stacking.
Like Van Halteren et al. \shortcite{vanhalteren98},
we have used different input versions:
one containing only the classifier output and another
containing both classifier output and a compressed representation
of the data item under consideration.
For the latter purpose we have used the part-of-speech tag of the
current word.

\section{Results$^4$}

\footnotetext[4]{Detailed results of our experiments are available on
http://lcg-www.uia.ac.be/\~{ }erikt/npcombi/}
\addtocounter{footnote}{1}

We want to find out whether system combination could
improve performance of baseNP recognition and, if this is the fact,
we want to select the best combination technique.
For this purpose we have performed an experiment with sections 15-18 of 
the WSJ part of the Penn Treebank as training data (211727 tokens) 
and section 21 as test data (40039 tokens).
Like the data used by Ramshaw and Marcus \shortcite{ramshaw95},
this data was retagged by the
Brill tagger in order to obtain realistic part-of-speech (POS) tags\footnote{
The retagging was necessary to assure that the performance rates 
obtained here would be similar to rates obtained for texts for which
no Treebank POS tags are available. 
}.
The data was segmented into baseNP parts and non-baseNP parts in a
similar fashion as the data used by Ramshaw and Marcus \shortcite{ramshaw95}.
Of the training data, only 90\% was used for training.
The remaining 10\% was used as tuning data for determining the weights
of the combination techniques.

For three classifiers (MBL, MaxEnt and IGTree) we have used
system-internal combination.
These learning algorithms have processed five different
representations of the output (IOB1, IOB2, IOE1, IOE2 and O+C) and
the results have been combined with majority voting.
The test data results can be found in Table \ref{tab-res-internal}.
In all cases, the combined results were better than that of the
best included system.

The results of ALLiS, {\sc c5.0}, MBSL and SNoW have been converted 
to the O and the C representation.
Together with the bracket representations of the other three techniques,
this gave us a total of seven O results and seven C results.
These two data streams have been combined with the combination
techniques described in section \ref{sec-combi}.
After this, we built baseNPs from the O and C results of each combination 
technique, like described in section \ref{sec-data-repr}.
The bracket accuracies and the F$_{\beta=1}$ scores for test data can
be found in Table \ref{tab-res1}. 

\begin{table}[t]
\begin{center}
\begin{tabular}{|l|c|c|c|}\cline{2-4}
\multicolumn{1}{l|}{section 21} & O & C & F$_{\beta=1}$ \\\hline
{\bf Classifier} &&&\\
ALLiS            & 97.87\% & 98.08\% & 92.15\\
{\sc c5.0}       & 97.05\% & 97.76\% & 89.97\\
IGTree           & 97.70\% & 97.99\% & 91.92\\
MaxEnt           & 97.94\% & 98.24\% & 92.60\\
MBL              & 98.04\% & 98.20\% & 92.82\\
MBSL             & 97.27\% & 97.66\% & 90.71\\
SNoW             & 97.78\% & 97.68\% & 91.87\\\hline
{\bf Simple Voting} &&&\\
Majority         & 98.08\% & 98.21\% & 92.95\\
TotPrecision     & 98.08\% & 98.21\% & 92.95\\
TagPrecision     & 98.08\% & 98.21\% & 92.95\\
Precision-Recall & 98.08\% & 98.21\% & 92.95\\\hline
{\bf Pairwise Voting} &&&\\
TagPair          & 98.13\% & 98.23\% & 93.07\\\hline
{\bf Memory-Based} &&&\\
Tags             & 98.24\% & 98.35\% & 93.39\\
Tags + POS       & 98.14\% & 98.33\% & 93.24\\\hline
{\bf Decision Trees} &&&\\
Tags             & 98.24\% & 98.35\% & 93.39\\
Tags + POS       & 98.13\% & 98.32\% & 93.21\\\hline
\end{tabular}
\end{center}
\caption{Bracket accuracies and F$_{\beta=1}$ scores for section WSJ 21 of
the Penn Treebank with seven individual classifiers and combinations of them.
Each combination performs better than its best individual member.
The stacked classifiers without context information perform best.
}
\label{tab-res1}
\end{table}

\begin{table*}[t]
\begin{center}
\begin{tabular}{|l|c|c|c|c|}\cline{2-5}
\multicolumn{1}{l|}{section 20}
                   & accuracy & precision & recall & F$_{\beta=1}$\\\hline
Best-five combination   & O:98.32\% C:98.41\% & 94.18\% & 93.55\% & 93.86 \\\hline
Tjong Kim Sang
\shortcite{tks2000a}    & O:98.10\% C:98.29\% & 93.63\% & 92.89\% & 93.26 \\
Mu\~noz et al.
\shortcite{munoz99}     & O:98.1\%  C:98.2\%  & 92.4\%  & 93.1\%  & 92.8  \\
Ramshaw and Marcus
\shortcite{ramshaw95}   & IOB1:97.37\%        & 91.80\% & 92.27\% & 92.03 \\
Argamon et al.
\shortcite{argamon99}   & -                   & 91.6\%  & 91.6\%  & 91.6  \\\hline
\end{tabular}
\end{center}
\caption{The overall performance of the majority voting combination of 
our best five systems (selected on tuning data performance) applied to 
the standard data set put forward by Ramshaw and Marcus 
\shortcite{ramshaw95} together with
an overview of earlier work.
The accuracy scores indicate how often a word was classified correctly
with the representation used (O, C or IOB1).
The combined system outperforms all earlier reported results for this 
data set.
}
\label{tab-res-tot}
\end{table*}

All combinations improve the results of the best individual
classifier.
The best results were obtained with a memory-based stacked classifier.
This is different from the combination results presented in 
Van Halteren et al. \shortcite{vanhalteren98}, 
in which pairwise voting performed best.
However, in their later work stacked classifiers outperform voting
methods as well \cite{vanhalteren00}.

Based on an earlier combination study \cite{tks2000a} 
we had expected the voting methods to do better.
We suspect that their performance is below that of the stacked 
classifiers because the difference between the best and the worst
individual system is larger than in our earlier study.
We assume that the voting methods might perform better if they were only 
applied to the classifiers that perform well on this task.
In order to test this hypothesis, we have repeated the combination 
experiments with the best n classifiers, where n took values from 3 to 
6 and the classifiers were ranked based on their performance on the 
tuning data.
The best performances were obtained with five classifiers:
F$_{\beta=1}$=93.44 for all five voting methods with the best
stacked classifier reaching 93.24.
With the top five classifiers, the voting methods outperform the best
combination with seven systems\footnote{
We are unaware of a good method for determining the significance of 
F$_{\beta=1}$ differences but we assume that 
this F$_{\beta=1}$ difference is not significant.
However, we believe that the fact that more combination methods perform 
well, shows that it easier to get a good performance out of the best
five systems than with all seven.}.
Adding extra classification results to a good combination system 
should not make overall performance worse so it is clear that there 
is some room left for improvement of our combination algorithms.

We conclude that the best results in this task can be obtained
with the simplest voting method, majority voting, applied to the
best five of our classifiers. 
Our next task was to apply the combination approach to a standard data
set so that we could compare our results with other work.
For this purpose we have used the data put forward by
Ramshaw and Marcus \shortcite{ramshaw95}.
Again, only 90\% of the training data was used for training while the
remaining 10\% was reserved for ranking the classifiers.
The seven learners were trained with the same parameters as in the
previous experiment.
Three of the classifiers (MBL, MaxEnt and IGTree) used
system-internal combination by processing different output
representations. 

The classifier output was converted to the O and the C representation.
Based on the tuning data performance, the classifiers ALLiS, 
{\sc igtree}, MaxEnt, MBL and SNoW were selected for being combined
with majority voting.
After this, the resulting O and C representations were combined to
baseNPs by using the method described in section \ref{sec-data-repr}.
The results can be found in Table \ref{tab-res-tot}.
Our combined system obtains an F$_{\beta=1}$ score of 93.86 which
corresponds to an 8\% error reduction compared with the best published
result for this data set (93.26). 

\section{Concluding remarks}

In this paper we have examined two methods for combining the results
of machine learning algorithms for identifying base noun phrases.
In the first method, the learner processed different output data
representations and the results were combined by majority voting.
This approach yielded better results than the best included
classifier.
In the second combination approach we have combined the results of
seven learning systems (ALLiS, {\sc c5.0}, IGTree, MaxEnt, MBL,
MBSL and SNoW). 
Here we have tested different combination methods.
Each combination method outperformed the best individual learning
algorithm and a majority vote of the top five systems performed best.
We have applied this approach of system-internal and system-external
combination to a standard data set for base noun phrase identification
and the performance of our system was better than any other published
result for this data set.

Our study shows that the combination methods that we have tested
are sensitive for the inclusion of classifier results of poor
quality.
This leaves room for improvement of our results by evaluating other 
combinators.
Another interesting approach which might lead to a better performance
is taking into account more context information, for example by
combining complete phrases instead of independent brackets.
It would also be worthwhile to evaluate using more elaborate methods 
for building baseNPs out of open and close bracket candidates.

\section*{Acknowledgements}

D\'ejean, Koeling and Tjong~Kim Sang are funded by the TMR
network Learning Computational 
Grammars\footnote{http://lcg-www.uia.ac.be/}. 
Punyakanok and Roth are supported by NFS grants IIS-9801638 and
SBR-9873450. 

\bibliographystyle{acl}

\end{document}